# Multimodal Machine Learning for Early Prediction of Metastasis in a Swedish Multi-Cancer Cohort

Franco Rugolon BSc, MMedSc, MSc[1]
Korbinian Randl BSc, MSc, MSc[1]
Braslav Jovanovic MD, PhD[2]
Ioanna Miliou MEng, PhD[1]
Panagiotis Papapetrou BSc, MA, PhD[1]

**Abstract**

**Purpose:** Multimodal Machine Learning offers a holistic view of a patient's status, integrating structured and unstructured data from electronic health records (EHR). We propose a framework to predict metastasis risk one month prior to diagnosis, using six months of clinical history from EHR data.

**Methods**: Data from four cancer cohorts collected at Karolinska University Hospital (Stockholm, Sweden) were analyzed: breast (n = 743), colon (n = 387), lung (n = 870), and prostate (n = 1890). The dataset included demographics, comorbidities, laboratory results, medications, and clinical text. We compared traditional and deep learning classifiers across single modalities and multimodal combinations, using various fusion strategies and a Transparent Reporting of a multivariable prediction model for Individual Prognosis Or Diagnosis (TRIPOD) 2a design, with an 80-20 development-validation split to ensure a rigorous, repeatable evaluation. Performance was evaluated using AUROC, AUPRC, F1 score, sensitivity, and specificity. We then employed a multimodal adaptation of SHAP to analyze the classifiers' reasoning.

**Results**: Intermediate fusion achieved the highest F1 scores on breast (0.845), colon (0.786), and prostate cancer (0.845), demonstrating strong predictive performance. For lung cancer, the intermediate fusion achieved an F1 score of 0.819, while the text-only model achieved the highest, with an F1 score of 0.829. Deep learning classifiers consistently outperformed traditional models. Colon cancer, the smallest cohort, had the lowest performance, highlighting the importance of sufficient training data. SHAP analysis showed that the relative importance of modalities varied across cancer types.

**Conclusion**: Fusion strategies offer distinct strengths and weaknesses. Intermediate fusion consistently delivered the best results, but strategy choices should align with data characteristics and organizational needs.

**Acknowledgements:** This work was partially funded by the Stiftelsen Marcus och Amalia Wallenbergs Minnesfond (The Marcus and Amalia Wallenberg Memorial Fund).

This work was partially funded by the Horizon Europe Research and Innovation programme under Grant Agreements No 875351 and 101093026.

**Corresponding author:** Franco Rugolon, **Affiliation:** Department of Computer and Systems Sciences, Stockholm University, Stockholm, Sweden. **Email address:** franco.rugolon@dsv.su.se.

**Running head:** Multimodal Machine Learning for Early Prediction of Metastasis.

**Context Summary:** *Key objective:* How accurately can multimodal machine learning predict the onset of metastasis one month in advance for patients of four major cancer cohorts, and which are the most influential features across several clinical data modalities contributing to the predictions?

*Knowledge generated:* Multimodal classifiers consistently outperformed unimodal approaches, achieving F1 scores above 81% across breast, lung, and prostate cancer cohorts, and an F1 score of 78% in the colon cancer cohort. A multimodal adaptation of SHAP demonstrated that the models dynamically weighted different modalities according to the cancer cohort, capturing cohort-specific prognostic information.

# Introduction

Cancer remains one of the leading causes of death worldwide despite advances in early diagnosis and treatment that have lowered mortality rates[1]. In 2025, Europe and the US are projected to experience approximately 6.5 million new cases and 2.6 million cancer-related deaths[1,2]. Breast, colon, lung, and prostate cancers have the highest incidence and mortality, accounting for nearly half of new cases and over one-third of US cancer deaths in 2024[3]. Despite therapeutic progress improving survival rates[4–6], significant clinical challenges persist.

Advanced-stage patients face high relapse risk through metastasis, resulting in poor prognosis. Metastasis is a complex process involving cancerous cells migrating, adapting, and establishing a vascular supply in new parts of the body[7]. Metastases increase tumor burden, reduce quality of life and decrease survival[7]. Reliable markers of metastatic spread remain elusive.

Machine Learning (ML) has shown promise in oncology by uncovering patterns in complex datasets, facilitating earlier diagnoses, effective treatments, and optimized resource allocation[8,9]. Electronic health records (EHRs) *structured* (e.g., demographics, comorbidities, and medications) and *unstructured* data (e.g., clinical notes and imaging). By integrating longitudinal and multifaceted health information, EHRs enhance clinical decision-making and provide a rich foundation for developing robust ML models[9–12]. Conventional ML approaches predominantly focused on data modalities in isolation, restricting their capacity to identify cross-modal relationships and the complexity of disease[13]. Multimodal ML (MML) enables the simultaneous analysis of multiple data modalities, including static, time-evolving, text, imaging, pathology, and omics, and can better replicate the integrative reasoning processes of clinicians. Recent studies have leveraged MML techniques by combining structured EHR data[14–21] with unstructured clinical notes[14], medical imaging[14,18,20,22–25], -omics[14,15,19,26], and pathology data[15–17,19,21,26] to support a range of oncology-related tasks, such as metastasis prediction[14,15,17–20,23–26], metastasis detection[27], and patient survival[21,23] in breast[14–16], colon[17,26,27], lung[18,23–25], and prostate[19–21] cancer. These studies often rely on advanced, non-routinely available data (e.g., -omics data) or data requiring expert labelling, such as medical imaging, to train their models.

The aim of this study is to use MML to predict the onset of metastasis one month in advance, based on routinely collected clinical data, while also providing insights into the underlying factors driving the model predictions. We implement three distinct multimodal fusion strategies alongside an MML-adapted explainability technique, enabling the extraction of relevant features across modalities that drive predictive performance. We focused on the most common cancers, i.e., breast, colon, lung, and prostate, using real-life data from patients treated at Karolinska University Hospital. This is a retrospective, single-institution study aimed to see whether this approach can be used as proof-of-concept of predicting imminent metastasis from routinely collected EHR data with MML. This research builds on our previous work[28], covering more patient cohorts and more sophisticated ML approaches.

# Methods

## The SU-ADE Corpus

We used the Stockholm EPR Structured and Unstructured ADE corpus (SU-ADE Corpus), extracted from the Swedish Health Records Research Bank[29][1], which contains data on more than 2 million patients treated at Karolinska University Hospital (Stockholm, Sweden), between 2006 and 2014, including ICD-10 diagnoses, ATC-coded medications, NPU-coded laboratory exams, and clinical notes (daily and discharge). We also included demographic data in the form of gender and year of birth.

## Cohort generation

We created four cancer patient cohorts, including patients with at least one occurrence of an ICD-10 diagnosis of malignant neoplasm of breast (C50), colon (C18), lung (C34), or prostate (C61)[30]. Metastatic onset was defined based on the ICD-10 codes C77-C79, the most reliable option available in the SU-ADE Corpus despite potential coding delays. For each patient, we extracted seven months of EHR data preceding either the initial recorded diagnosis of metastasis or, if absent, the most recent clinical event. To ensure a clinically meaningful prediction window, we restricted inclusion to patients with at least seven months of clinical history. Patients with shorter records (e.g., <3 months) often represent late palliative cases with very limited survival and were therefore excluded to avoid confounding the analysis. This ensured sufficient follow-up (≥6 months), which is routinely considered clinically informative[31]. For model development, only data from the first six months were utilized, with the final month excluded to establish a clinically actionable prediction window.

---

[1] This research has been approved by the Regional Ethical Review Board in Stockholm under permission no. 2014/1882-31/5.

*Table 1: A table presenting the distribution of the patients having breast, colon, lung, and prostate cancer into age groups, gender groups, and specifying the location of their metastasis (if any).*

|  |  |  | Overall | Metastasis Yes | Metastasis No |
|---|---|---|---|---|---|
| **Breast Cancer (C50)** | n |  | 743 | 281 | 462 |
|  | Age Group, n (%) | Youth (<25 years) | 1 (0.1) | 1 (0.4) |  |
|  |  | Adults (25-64 years) | 380 (51.1) | 160 (56.9) | 220 (47.6) |
|  |  | Seniors (>64 years) | 362 (48.7) | 120 (42.7) | 242 (52.4) |
|  | Gender, n (%) | Female | 734 (98.8) | 275 (97.9) | 459 (99.4) |
|  |  | Male | 9 (1.2) | 6 (2.1) | 3 (0.6) |
|  | Metastasis location, n (%) | Lymph Nodes |  | 102 (36.3) |  |
|  |  | Other and Unspecified Sites |  | 59 (21.0) |  |
|  |  | Respiratory and Digestive Organs |  | 120 (42.7) |  |
| **Colon Cancer (C18)** | n |  | 387 | 111 | 276 |
|  | Age Group, n (%) | Youth (<25 years) | 6 (1.6) | 1 (0.9) | 5 (1.8) |
|  |  | Adults (25-64 years) | 126 (32.6) | 44 (39.6) | 82 (29.7) |
|  |  | Seniors (>64 years) | 255 (65.9) | 66 (59.5) | 189 (68.5) |
|  | Gender, n (%) | Female | 192 (49.6) | 51 (45.9) | 141 (51.1) |
|  |  | Male | 194 (50.1) | 60 (54.1) | 134 (48.6) |
|  |  | Other | 1 (0.3) |  | 1 (0.4) |
|  | Metastasis location, n (%) | Lymph Nodes |  | 31 (27.9) |  |
|  |  | Other and Unspecified Sites |  | 7 (6.3) |  |
|  |  | Respiratory and Digestive Organs |  | 73 (65.8) |  |
| **Lung Cancer (C34)** | n |  | 870 | 458 | 412 |
|  | Age Group, n (%) | Youth (<25 years) | 1 (0.1) | 1 (0.2) |  |
|  |  | Adults (25-64 years) | 230 (26.4) | 150 (32.8) | 80 (19.4) |
|  |  | Seniors (>64 years) | 639 (73.4) | 307 (67.0) | 332 (80.6) |
|  | Gender, n (%) | Female | 473 (54.4) | 256 (55.9) | 217 (52.7) |
|  |  | Male | 397 (45.6) | 202 (44.1) | 195 (47.3) |
|  | Metastasis location, n (%) | Lymph Nodes |  | 100 (21.8) |  |
|  |  | Other and Unspecified Sites |  | 181 (39.5) |  |
|  |  | Respiratory and Digestive Organs |  | 177 (38.6) |  |
| **Prostate Cancer (C61)** | n |  | 1890 | 515 | 1375 |
|  | Age Group, n (%) | Adults (25-64 years) | 452 (23.9) | 104 (20.2) | 348 (25.3) |
|  |  | Seniors (>64 years) | 1438 (76.1) | 411 (79.8) | 1027 (74.7) |
|  | Gender, n (%) | Male | 1890 (100.0) | 515 (100.0) | 1375 (100.0) |
|  | Metastasis location, n (%) | Lymph Nodes |  | 183 (35.5) |  |
|  |  | Other and Unspecified Sites |  | 244 (47.4) |  |

| | | |
|---|---|---|
| Respiratory and Digestive Organs | | 88 (17.1) |

We included only patients with all modalities available (demographic information, comorbidities, laboratory exams, medications, and clinical notes) and whose clinical history was at least seven months long between the first cancer diagnosis and the last recorded event, ensuring active cancer care for the included patients. Patients who did not survive through the seven-month observation period were excluded. Patients who died after a metastasis diagnosis were labeled positive; those who died without a metastasis code were labeled negative.

The final cohorts comprised 743 breast, 387 colon, 870 lung, and 1,890 prostate cancer patients (Table 1). Imbalance varied by cohort: lung cancer was most balanced (ratio 0.9), followed by breast (0.6), with colon (0.4) and prostate (0.38) more imbalanced. Furthermore, we extracted and preprocessed both structured and unstructured data sources:

- **Demographics.** Age was discretized into three groups: "youth" (<25 years), "adult" (25–64 years), and "senior" (>64 years), based on the age at the beginning of the observation period. Both age and gender were converted into binary indicators using one-hot encoding.

- **Comorbidities.** Diagnoses recorded during the six-month observation period were one-hot encoded. We exploited the hierarchical structure of ICD-10 codes to aggregate non-cancer diagnoses by their first three digits, without losing any level of detail for the neoplasm-specific diagnoses.

- **Laboratory exams.** Lab results were modeled as monthly time series. For each exam, we computed the monthly average of the values; exams that were never performed in a given month were assigned a value of -1.

- **Medications.** Medication prescriptions were aggregated monthly by counting the occurrences of each medication per month, and grouped into the ATC pharmacological subgroups.

- **Clinical notes.** Clinical notes were represented as event sequences. To maintain relevance and computational efficiency, we selected the 20 most recent clinical notes within the six-month window, ensuring the model received concise yet informative input while minimizing hardware resource demands. The text was further tokenized using a SweDeClinBERT WordPiece tokenizer developed for Swedish clinical text[32].

After preprocessing, four data modalities were defined: **Static**, variables that do not change over time (demographics and comorbidities); **Labs** (time-stamped laboratory results); **Meds** (drug prescriptions over time); and **Text** (clinical notes). Static, Meds, and Labs represent structured data, while Text represents unstructured data.

## Multimodal classification

The multimodal classification task involved integrating four modalities to identify reliable markers of metastatic onset in patients with breast, colon, lung, or prostate cancer.

### Unimodal Baselines.

We trained a set of *unimodal classifiers* selected according to the underlying data type of each modality, as detailed below:

- **Static (Tabular).** We compared three traditional ML classifiers: K-Nearest Neighbors (KNN)[33], Logistic Regression (LR)[34], Gradient Boosting (GB)[35], and one Deep Learning (DL) based classifier, a Multilayer Perceptron (MLP) with one hidden layer, batch normalization, dropout, and a sigmoid output. The hidden layer used ReLU activation.

- **Meds and Labs (Time series).** We used two traditional ML classifiers: a variation of KNN adapted to time series, Catch22[36], and a DL classifier, a Recurrent Neural Network (RNN) with Gated Recurrent Units (GRU) with a similar architecture to the MLP but with tanh hidden activation.

- **Text (Event sequences).** We trained a DL classifier composed of a pretrained encoder-only transformer[37], followed by a GRU layer, batch normalization, dropout, and a sigmoid output, with tanh hidden activation.

### Multimodal approaches.

We implemented three multimodal fusion strategies, as shown in Figure 1.

- **Early Fusion (EF).** Structured modalities (Static, Meds, and Labs) were converted into a unified tabular format by averaging the time series values and then concatenating them with the static features. The same classifiers used for Static data were trained on the fused data. Text was excluded due to its unstructured nature.

- **Late Fusion (LF).** Predictions from the best unimodal classifiers were combined using a weighted average ensemble with weights proportional to mean validation AUPRCs.

- **Intermediate Fusion (IF).** The best unimodal DL classifiers were selected based on their validation performance. By removing the output layer from each classifier, they were repurposed as encoders to produce individual lower-dimensional latent representations of the unimodal data, which were then concatenated into a unified latent space, followed by an additional layer with ReLU activation, batch normalization, dropout, and a sigmoid output. Initially, only the new layers were trained while encoder weights were frozen, followed by end-to-end fine-tuning.

More details on the hyperparameter ranges and training sequence can be found in Appendix: Validation details.

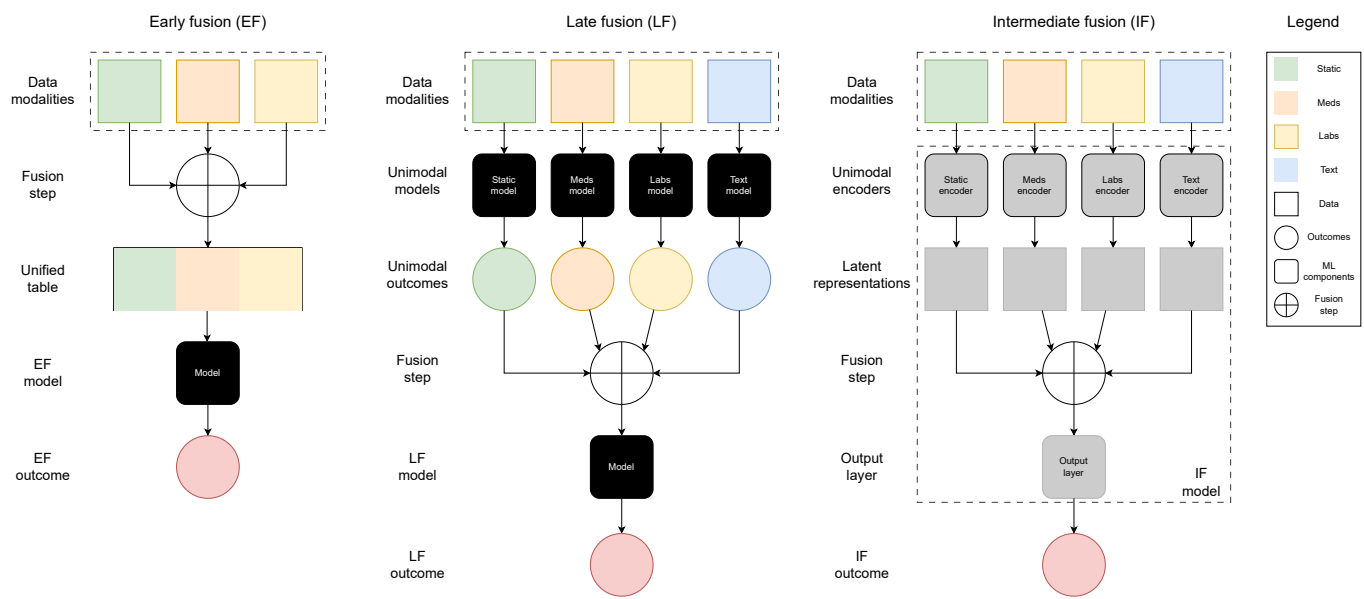

*Figure 1: A graphical representation of the various fusion strategies.*

## Validation, testing, and implementation

All classifiers were trained and evaluated following a TRIPOD Type 2a design[38], using a single cohort that was randomly split into a development set (80%) and an internal validation set (20%), with stratification to preserve outcome prevalence. Model training and hyperparameter tuning were performed exclusively on the development set, using 5-fold stratified cross-validation. Model selection was based on maximizing the area under the precision–recall curve (AUPRC) to account for class imbalance.

Final model performance was assessed on the held-out internal validation set, from which we computed the AUPRC, AUROC, F1 score, sensitivity, and specificity. Hyperparameter ranges and full training details are reported in Appendix: Validation details.

As an additional robustness analysis, we conducted a nested stratified cross-validation, which provides performance estimates across multiple train/test splits, and supports more stable comparisons between classifiers[39,40]. The design and results are reported in Appendix: Nested Stratified cross-validation analysis.

All experiments were conducted using Python 3.10.12. Scikit-learn[41] was used to implement traditional models, sktime[42] for time series models, and Keras[43] with a PyTorch[44] backend for deep learning models.

## Explainability

To interpret the decision-making process of the best-performing models, we employed SHAP (SHapley Additive exPlanations)[45], which approximates Shapley values to estimate the impact of model input on the output[46]. Given a black-box model and a specific input, SHAP highlights the most influential components of the input by selectively hiding parts of it and measuring how the predictions change. While SHAP is traditionally designed for unimodal classifiers, we adapted it for multimodal architectures, as shown in Appendix: Serialization of multimodal data for use with SHAP. Each timestep in the time series and each token position from the clinical notes was represented as a distinct feature, and a distance metric captured relative positions across tokens or timesteps.

To validate our approach, we investigated the behaviour of the classifiers during the SHAP analysis, and the results can be seen in Appendix: Analysis of SHAP faithfulness.

# Results

Experimental results on multimodal classification for all cohorts are summarized in Table 2**Error! Reference source not found.**.

## Comparison by modality.

In accordance with its widespread adoption in literature[15–18,20,23,24,26,27], IF was the best way of combining modalities, surpassing other fusion strategies and single-modality classifiers, achieving the best AUPRC and F1 scores in the breast, colon, and prostate cohorts, while SweDeClinBERT was the best classifier in lung cancer. The F1 score for IF exceeded 0.80 for all cohorts except colon cancer, indicating strong predictive performance. It also had the highest specificity in all cohorts except for lung cancer, where SweDeClinBERT achieved the highest specificity.

Among the unimodal classifiers, the one trained on Text was the top performer, with F1 scores over 0.72 across all cohorts, likely due to the effectiveness of the pretrained SweDeClinBERT model.

The EF model achieved performance comparable to the unimodal classifiers for structured data, but generally it did not surpass them. We attribute this to the necessity of bringing all modalities to a common feature space (e.g., by averaging over time), which may lead to the loss of important temporal trends.

Finally, the LF model outperformed the results of the classifiers deployed on the individual structured modalities in all cohorts, achieving results close to the text classifiers, as shown in Table 2. In the colon cancer cohort, the results for LF were similar to those of EF. Additionally, in all cohorts, deep learning models consistently outperformed classical models and time series models, both in all single modalities and EF.

## Comparison by cancer type.

Colon cancer exhibited the lowest predictive performance across all modalities and fusion strategies, with an F1 score of 0.78 for IF, with the overall performance likely being influenced by the smaller sample size of 387 patients (see Table 1), which may have hindered the generalization capabilities of the classifiers.

For prostate cancer, the RNN trained on Meds achieved the highest specificity but suffered from low sensitivity. The IF (censored) classifier performed equally well as the IF model in breast and lung cancer, which can be explained by the potential lack of censored sentences in the specific internal validation split. Despite the class imbalance in the breast and prostate cohorts, with positive class proportions of 37.82% and 27.25%, respectively, these cohorts achieved higher performance, suggesting that the imbalance was not a major limiting factor.

Compared to prior studies on metastasis prediction, our IF classifiers demonstrate competitive, and in some case superior, performance compared to the current state of the art, depending on cancer type and evaluation setting. For **breast cancer**, previous studies reported an AUROC of 0.75[15], whereas our IF model achieved a substantially higher AUROC of 0.837 in our cohort. Similarly, while a previous study reported an F1 score of 0.82[14], our model improved upon this with an F1 score of 0.845. For **lung cancer,** published AUROC values include 0.762 for an ensemble model and 0.756 for the best single model under internal validation[18], as well as a range from 0.820 to 0.865 under internal validation in a separate study[23]. In our lung cancer cohort, the IF model achieved an AUROC of 0.819, and the text model, the best performing approach in that cohort, reached an AUROC of 0.829.  For **prostate cancer**, a reported AUROC of 0.89[20] compares with an AUROC of 0.861 achieved by our IF model. Notably, our results are obtained using only routinely collected clinical data, in contrast to many prior studies that rely on pathology or imaging data. Instances of lower performance may reflect differences in data availability, such as the inclusion of imaging or pathology, or the use of larger patient cohorts in other studies.

Table 2: A summary of the results, showing the best classifier for each modality or combination of modalities for breast, colon, lung, and prostate cancer, respectively, presenting the value computed on the held-out test for each metric. The best result for each metric for each cancer is presented in **bold**.

| | Modality | Classifier | AUPRC | AUROC | F1 Score (macro) | Specificity | Sensitivity |
|---|---|---|---|---|---|---|---|
| Breast | Static | MLP | 0.508 | 0.677 | 0.677 | 0.639 | 0.747 |
| | Labs | RNN | 0.547 | 0.685 | 0.693 | 0.481 | 0.889 |
| | Meds | RNN | 0.437 | 0.581 | 0.563 | 0.264 | 0.905 |
| | Text | SweDeClinBERT | 0.665 | 0.804 | 0.799 | 0.771 | 0.846 |
| | EF (w/o text) | MLP | 0.533 | 0.671 | 0.678 | 0.455 | 0.896 |
| | LF | Avg (weighted) | 0.706 | 0.811 | 0.822 | 0.678 | 0.952 |
| | IF | IF | **0.766** | **0.837** | **0.845** | **0.694** | **0.994** |
| | IF (censored) | IF | **0.766** | **0.837** | **0.845** | **0.694** | **0.994** |
| Colon | Static | MLP | 0.274 | 0.540 | 0.538 | 0.326 | 0.766 |
| | Labs | RNN | 0.312 | 0.542 | 0.508 | 0.095 | 0.984 |
| | Meds | RNN | 0.291 | 0.530 | 0.524 | 0.019 | 0.907 |
| | Text | SweDeClinBERT | 0.483 | 0.704 | 0.723 | 0.473 | 0.925 |
| | EF (w/o text) | MLP | 0.319 | 0.609 | 0.566 | 0.642 | 0.583 |
| | LF | Avg (weighted) | 0.348 | 0.593 | 0.597 | 0.244 | **0.927** |
| | IF | IF | **0.540** | **0.820** | **0.786** | **0.806** | 0.848 |
| | IF (censored) | IF | 0.451 | 0.729 | 0.715 | 0.679 | 0.848 |
| Lung | Static | MLP | 0.602 | 0.635 | 0.635 | 0.649 | 0.611 |
| | Labs | RNN | 0.629 | 0.655 | 0.652 | 0.776 | 0.520 |
| | Meds | RNN | 0.590 | 0.603 | 0.603 | 0.593 | 0.615 |
| | Text | SweDeClinBERT | **0.795** | **0.829** | **0.832** | **0.879** | 0.775 |
| | EF (w/o text) | MLP | 0.586 | 0.583 | 0.582 | 0.637 | 0.544 |
| | LF | Avg (weighted) | 0.762 | 0.793 | 0.794 | 0.870 | **0.816** |
| | IF | IF | 0.777 | 0.819 | 0.819 | 0.848 | 0.795 |
| | IF (censored) | IF | 0.777 | 0.819 | 0.819 | 0.848 | 0.795 |
| Prostate | Static | MLP | 0.352 | 0.622 | 0.629 | 0.365 | 0.881 |
| | Labs | RNN | 0.518 | 0.715 | 0.745 | 0.468 | **0.965** |
| | Meds | RNN | 0.375 | 0.614 | 0.623 | 0.277 | 0.950 |
| | Text | SweDeClinBERT | 0.593 | 0.791 | 0.807 | 0.650 | 0.935 |
| | EF (w/o text) | MLP | 0.400 | 0.686 | 0.679 | 0.546 | 0.809 |
| | LF | Avg (weighted) | **0.643** | 0.793 | 0.828 | 0.599 | 0.982 |
| | IF | IF | 0.605 | **0.861** | **0.845** | **0.800** | 0.946 |
| | IF (censored) | IF | 0.602 | 0.848 | 0.841 | 0.762 | 0.950 |

## Explainability.

Following model evaluation, we applied SHAP to explain the predictions of the IF models, which achieved the highest overall performance in the breast, colon, and prostate cohorts. Although the text-only model marginally outperformed IF in the lung cancer cohort, it is inherently unimodal and therefore cannot support comparative analysis of cross-modal contributions. To enable consistent multimodal interpretability across all cancer types, we conducted SHAP analyses on the IF models in all four cohorts. We conducted both *global and local evaluations*, averaging importance scores across patients and per patient, respectively.

Globally (Figure 2), Text was the most relevant modality for colon and lung cancer, Static for breast cancer, and Meds for prostate cancer, with the latter two showing more balanced modality contributions overall. Labs was the least important modality across cancers except for prostate cancer, where Static ranked lowest. These findings suggest that modality importance varies across cancer types and clinical contexts.

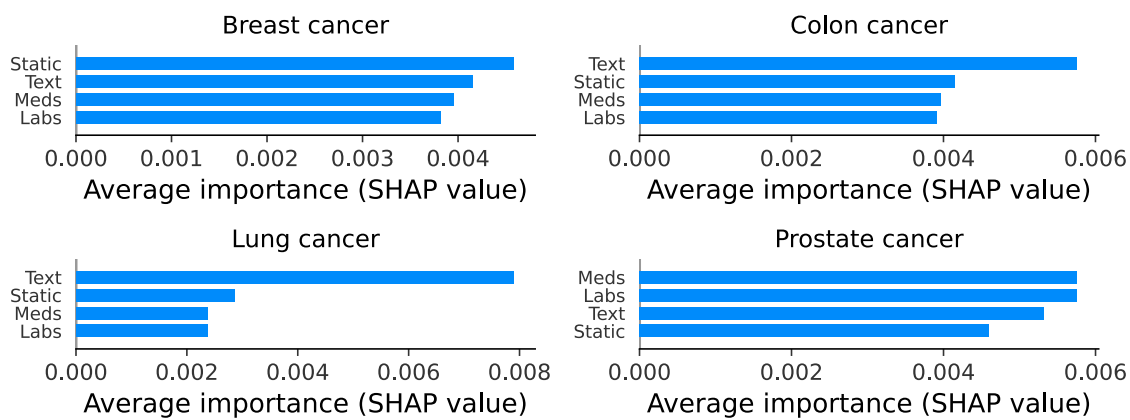

*Figure 2: Global relevance of the four modalities towards the prediction according to SHAP, for all four cancers.*

To gain deeper insight, we conducted local SHAP analysis focusing on features within the top 0.1 percentile of importance scores for each cancer cohort. We noted that individual feature relevance varied notably between patients, despite relatively uniform global modality contributions. This suggests that certain features are highly influential within specific clinical contexts, despite not showing global correlations across the entire cohort.

To further assess the robustness of the text modality and exclude reliance on explicit mentions of metastasis, we conducted a censoring analysis, detailed in Appendix: Evaluation of Model Robustness After Text Censoring.

### Limitations.

This was a retrospective proof-of-concept study based on data from a single institution. Our aim was to evaluate the feasibility of predicting imminent metastasis using routinely collected clinical data, rather than to establish a fully generalizable model. The study did not explore temporal or disease stage stratification, nor cross-cohort validation, which could provide additional insights into model robustness. Validation in external cohorts and in prospective settings will therefore be essential to confirm the translational potential of this approach.

The SU-ADE dataset was not originally designed for cancer-related research, and metastatic onset had to be defined using ICD10-based clinical codes, which may be affected by misclassification, coding delays, or retrospective documentation. Although more reliable than text-based alternatives, this strategy could be improved by integrating codes with temporal information from clinical notes or imaging reports.

We also employed a convenience sample, including patients with at least seven months of clinical history. While this criterion was chosen to reduce inclusion of late palliative trajectories, it may introduce selection bias by excluding shorter observation periods, and it may mix heterogeneous disease stages. Future studies should assess the impact of alternative cutoffs on model performance and generalizability.

Finally, this study forms part of a broader line of research in which we are extending the approach to cross-cohort validation and transfer learning, aiming to leverage data from common cancers to improve prediction in rarer malignancies such as sarcoma and central nervous system tumors, where sample sizes are limited.

## Conclusion

In this retrospective study, we developed and evaluated ML models for predicting metastasis in the four most prevalent human cancers using multimodal EHR data. We compared three multimodal fusion strategies, EF, LF, and IF, each with distinct strengths and limitations. While EF preserves cross-modal interactions, it may dilute modality-specific information and cannot accommodate all data types. LF preserves modality information but lacks the ability to exploit inter-modality relationships. IF provides a balanced approach, capturing both within- and cross-modality patterns, though it requires more complex architectures. Our results highlight clinical Text as the most informative modality when used in isolation, even though it introduces challenges related to its unstructured nature and lack of standardization.

These models could be applied at clinics, providing longitudinal risk monitoring by updating predictions based on the most recent patient history. Since the model operates solely in inference mode, it has low computational requirements and delivers results rapidly, making integration into existing clinical environment feasible. High risk patients could then be prioritized for additional evaluations enabling earlier detection of metastasis, while incurring no additional cost for routine use and concentrating resource-intensive assessments only on at-risk patients.

Appendix: Validation details

*Table 3: Hyperparameters for all the classifiers, separated by modality and depending on whether they are traditional or DL ones.*

| Category | Type | Classifier | Hyperparameters (search grid) |
|---|---|---|---|
| Static and EF | Traditional | GradientBoostingClassifier | n_estimators: 100, 200, 300; max_depth: 2, 3, 5, 10; random_state: 0 |
| | | RandomForestClassifier | n_estimators: 100, 200, 300; max_depth: 2, 3, 5, 10; random_state: 0 |
| | | LogisticRegression | C: 0.1, 1, 10; penalty: l1, l2; solver: liblinear; random_state: 0 |
| | DL | NN | Dropout: 0.2, 0.3; Units_multiplier: 1, 2, 3; Hidden_activation: ReLU |
| Labs and Meds | Traditional | RocketClassifier | num_kernels: 1000, 5000, 10000; random_state: 0 |
| | | MUSE | support_probabilities: True; anova: True, False; bigrams: True, False |
| | | Catch22Classifier | estimator: RandomForestClassifier (200 estimators), GradientBoostingClassifier (200 estimators), LogisticRegression; random_state: 0 |
| | DL | RNN | Dropout: 0.2, 0.3; Units_multiplier: 1, 2, 3; recurrent_layer_type: GRU; Hidden_activation: tanh |
| Text | DL | SweDeClinBERT | Dropout: 0.2, 0.3; Units_multiplier: 1 |
| LF | Traditional | Avg *(weighted)* | Weights derived from validation performance (per fold) |
| IF | DL | IF | Dropout: 0.2, 0.3. The best unimodal DL encoders were selected for each modality, initially using frozen weights, with a final fine-tuning passage. |

All models were developed and internally validated according to a TRIPOD Type 2a design. For every classifier, we performed an exhaustive search across all hyperparameter combinations in the validation folds of cross-validation. Each model–hyperparameter configuration was trained on the training folds and evaluated on the corresponding validation folds. All the models for each modality or combination of modalities and their corresponding hyperparameters can be seen in Table 3.

Training proceeded in three sequential stages:

- **Unimodal models and EF:** Each modality (Static, Labs, Medications, Text, etc.) was modeled independently. For each modality, the best-performing classifier and hyperparameter configuration were identified based on the mean validation AUPRC. The selected classifiers were then retrained on the entire training and validation folds.

- **LF:** The best performing classifiers were selected from each unimodal experiment, and their predictions were combined using a weighted average, where the weights were defined based on the validation performance of each model.
- **IF:** The best-performing deep learning–based classifiers from each unimodal experiment were adapted into modality-specific encoders by removing their final classification layers. Encoder outputs were concatenated, followed by a dense layer with ReLU activation, batch normalization, and dropout, and a final sigmoid output layer. Training was performed in two stages: first, the added layers were trained with encoder weights frozen; then all layers were unfrozen, and the full model was fine-tuned on the training+validation folds.

After the training procedure was completed, the final performance of the best-performing models was computed on the held-out internal validation set (see Table 2**Error! Reference source not found.**).

## Appendix: Nested Stratified cross-validation analysis

In addition to the TRIPOD type 2a study design, we also evaluated the models using nested, stratified 5-fold cross-validation. In this design, two cross-validation loops are implemented: an outer cross-validation loop, used to generate independent test folds, and an inner loop for model selection and hyperparameter tuning [39,40]. Each fold in the outer loop consisted of 20% of the original patients, while the remaining 80% of the patients was split by the internal loop in 64% training and 16% validation. For each outer iteration, one fold was held out for testing, and the remaining data were further split into training and validation folds within the inner loop. This procedure ensures that test data are not used during model selection, thereby reducing optimistic bias [40], and it is equivalent to repeating a single development–validation split, as defined in the TRIPOD type 2a design, 5 times, providing more stable performance estimates by averaging results across multiple train–test partitions and making more efficient use of limited data, at the cost of increased computational complexity. Performance was averaged across the folds of the internal loop, and the configuration with the highest average AUPRC was selected and tested on the outer folds.

To further increase the stability of the performance estimates, we also compute our performance metrics on bootstrapping samples from the test set, obtaining confidence intervals, which we report in Table 4.

*Table 4: A summary of the results computed using nested stratified cross validation and bootstrapping, showing the best classifier for each modality or combination of modalities for breast, colon, lung, and prostate cancer, respectively, presenting the value computed on the*

pooled test fold samples and 95% confidence intervals as Value, 95% CI [LL, UL] for each metric. The best result for each metric for each cancer is presented in **bold**.

| Modality | Classifier | AUPRC | AUROC | F1 Score (macro) | Specificity | Sensitivity |
|---|---|---|---|---|---|---|
| Static | MLP | 0.497 [0.360,0.609] | 0.649 [0.525, 0.732] | 0.614 [0.513, 0.733] | 0.493 [0.250, 0.704] | 0.811 [0.495, 0.928] |
| Labs | RNN | 0.525 [0.423, 0.675] | 0.661 [0.580, 0.767] | 0.607 [0.561, 0.778] | 0.426 [0.210, 0.656] | 0.909 [0.755, 0.989] |
| Meds | RNN | 0.423 [0.331, 0.547] | 0.538 [0.489, 0.659] | 0.376 [0.377, 0.653] | 0.104 [0.000, 0.440] | **0.977 [0.840, 1.000]** |
| Text | SweDeClinBERT | 0.676 [0.511, 0.847] | 0.818 [0.642, 0.916] | 0.818 [0.629, 0.909] | 0.805 [0.332, 0.949] | 0.855 [0.702, 0.978] |
| EF (w/o text) | MLP | 0.516 [0.384, 0.634] | 0.670 [0.554, 0.758] | 0.648 [0.549, 0.761] | 0.536 [0.274, 0.724] | 0.833 [0.681, 0.937] |
| LF | Avg (weighted) | 0.680 [0.546, 0.807] | 0.780 [0.644, 0.876] | 0.746 [0.628, 0.884] | 0.622 [0.300, 0.824] | 0.950 [0.851, 1.000] |
| IF | IF | **0.735 [0.562, 0.845]** | **0.861 [0.751, 0.943]** | **0.834 [0.750, 0.938]** | **0.785 [0.586, 0.999]** | 0.923 [0.813, 1.000] |
| IF (censored) | IF | 0.720 [0.546, 0.821] | 0.852 [0.747, 0.929] | 0.821 [0.744, 0.922] | 0.763 [0.535, 0.972] | 0.925 [0.810, 1.000] |
| Static | MLP | 0.293 [0.196, 0.477] | 0.514 [0.420, 0.657] | 0.439 [0.379, 0.660] | 0.176 [0.000, 0.571] | 0.857 [0.667, 0.981] |
| Labs | RNN | 0.334 [0.195, 0.454] | 0.522 [0.446, 0.634] | 0.458 [0.381, 0.651] | 0.050 [0.000, 0.300] | 0.999 [0.862, 1.000] |
| Meds | RNN | 0.285 [0.194, 0.447] | 0.499 [0.480, 0.583] | 0.446 [0.389, 0.568] | 0.000 [0.000, 0.215] | **1.000 [0.903, 1.000]** |
| Text | SweDeClinBERT | 0.440 [0.269, 0.713] | 0.655 [0.531, 0.818] | 0.545 [0.505, 0.819] | 0.401 [0.105, 0.704] | 0.934 [0.807, 1.000] |
| EF (w/o text) | MLP | 0.326 [0.198, 0.477] | 0.546 [0.440, 0.679] | 0.454 [0.402, 0.681] | 0.247 [0.044, 0.667] | 0.842 [0.540, 1.000] |
| LF | Avg (weighted) | 0.321 [0.195, 0.489] | 0.527 [0.492, 0.658] | 0.458 [0.394, 0.681] | 0.048 [0.000, 0.333] | **1.000 [0.919, 1.000]** |
| IF | IF | **0.509 [0.307, 0.700]** | **0.752 [0.609, 0.858]** | **0.633 [0.587, 0.865]** | **0.580 [0.183, 0.870]** | 0.946 [0.674, 1.000] |
| IF (censored) | IF | 0.491 [0.307, 0.681] | 0.735 [0.584, 0.842] | 0.613 [0.558, 0.857] | 0.532 [0.164, 0.774] | 0.957 [0.674, 1.000] |
| Static | MLP | 0.616 [0.527, 0.710] | 0.633 [0.529, 0.719] | 0.630 [0.524, 0.718] | 0.697 [0.470, 0.831] | 0.571 [0.392, 0.816] |
| Labs | RNN | 0.624 [0.540, 0.699] | 0.637 [0.532, 0.727] | 0.656 [0.482, 0.724] | 0.668 [0.383, 0.965] | 0.646 [0.171, 0.829] |
| Meds | RNN | 0.625 [0.525, 0.716] | 0.647 [0.539, 0.718] | 0.633 [0.518, 0.718] | 0.723 [0.408, 0.957] | 0.554 [0.213, 0.835] |
| Text | SweDeClinBERT | 0.763 [0.677, 0.845] | 0.805 [0.704, 0.874] | 0.797 [0.701, 0.873] | 0.847 [0.725, 0.947] | 0.753 [0.575, 0.911] |
| EF (w/o text) | MLP | 0.612 [0.514, 0.691] | 0.622 [0.507, 0.720] | 0.631 [0.448, 0.718] | 0.645 [0.204, 0.803] | 0.618 [0.430, 0.849] |
| LF | Avg (weighted) | 0.748 [0.656, 0.817] | 0.786 [0.697, 0.855] | 0.784 [0.700, 0.856] | 0.850 [0.724, 0.947] | 0.724 [0.569, 0.854] |
| IF | IF | **0.780 [0.670, 0.876]** | 0.815 [0.713, 0.895] | 0.822 [0.707, 0.894] | **0.855 [0.677, 0.977]** | 0.793 [0.560, 0.938] |
| IF (censored) | IF | **0.782 [0.685, 0.874]** | **0.819 [0.722, 0.901]** | **0.824 [0.718, 0.897]** | 0.854 [0.665, 0.989] | **0.798 [0.560, 0.962]** |
| Static | MLP | 0.393 [0.284, 0.485] | 0.647 [0.550, 0.725] | 0.542 [0.548, 0.729] | 0.418 [0.216, 0.597] | 0.878 [0.777, 0.948] |
| Labs | RNN | 0.491 [0.391, 0.608] | 0.701 [0.606, 0.782] | 0.592 [0.608, 0.803] | 0.462 [0.233, 0.628] | 0.946 [0.880, 1.000] |
| Meds | RNN | 0.346 [0.268, 0.474] | 0.565 [0.501, 0.662] | 0.364 [0.430, 0.680] | 0.155 [0.009, 0.399] | **0.977 [0.907, 1.000]** |
| Text | SweDeClinBERT | 0.641 [0.478, 0.766] | 0.817 [0.657, 0.904] | 0.747 [0.675, 0.896] | 0.694 [0.321, 0.921] | 0.946 [0.855, 0.993] |
| EF (w/o text) | MLP | 0.436 [0.309, 0.553] | 0.688 [0.602, 0.773] | 0.603 [0.608, 0.777] | 0.514 [0.280, 0.678] | 0.867 [0.731, 0.952] |
| LF | Avg (weighted) | 0.599 [0.419, 0.762] | 0.750 [0.597, 0.873] | 0.649 [0.590, 0.884] | 0.533 [0.194, 0.787] | 0.975 [0.941, 1.000] |
| IF | IF | **0.658 [0.503, 0.788]** | **0.864 [0.742, 0.940]** | **0.863 [0.751, 0.935]** | **0.755 [0.529, 0.935]** | 0.964 [0.862, 1.000] |

| IF (censored) | IF | 0.626 [0.476, 0.767] | 0.838 [0.725, 0.921] | 0.844 [0.739, 0.920] | 0.698 [0.469, 0.916] | 0.976 [0.870, 1.000] |

To assess whether performance differences among the best classifiers for each modality were statistically significant, we first applied Friedman's chi-square test separately within each cancer cohort on the performance calculated using nested stratified cross validation and bootstrapping, as explained in Appendix: Nested Stratified cross-validation analysis.

In all cohorts, the Friedman test yielded p ≤ 0.05, as can be seen in Table 5, indicating significant overall differences in performance.

*Table 5: Results of Friedman's chi-square tests evaluating whether differences in AUPRC performance across modalities and fusion strategies were statistically significant in each cancer cohort. Reported values are the test statistic ($\chi^2$) and corresponding p-value.*

| **Cohort** | Test statistic ($\chi^2$) | p-value |
| --- | --- | --- |
| Breast cancer | 148.91 | $6.86*10^{-29}$ |
| Colon cancer | 124.76 | $7.79*10^{-24}$ |
| Lung cancer | 139.53 | $6.38*10^{-27}$ |
| Prostate cancer | 141.73 | $2.20*10^{-27}$ |

When the Friedman test indicated overall significance, we conducted Nemenyi post-hoc pairwise comparisons, as implemed in scikit_posthocs, to identify which modalities or fusion strategies had significantly different performances from one another. The results of these comparisons, can be seen in Figure 3.

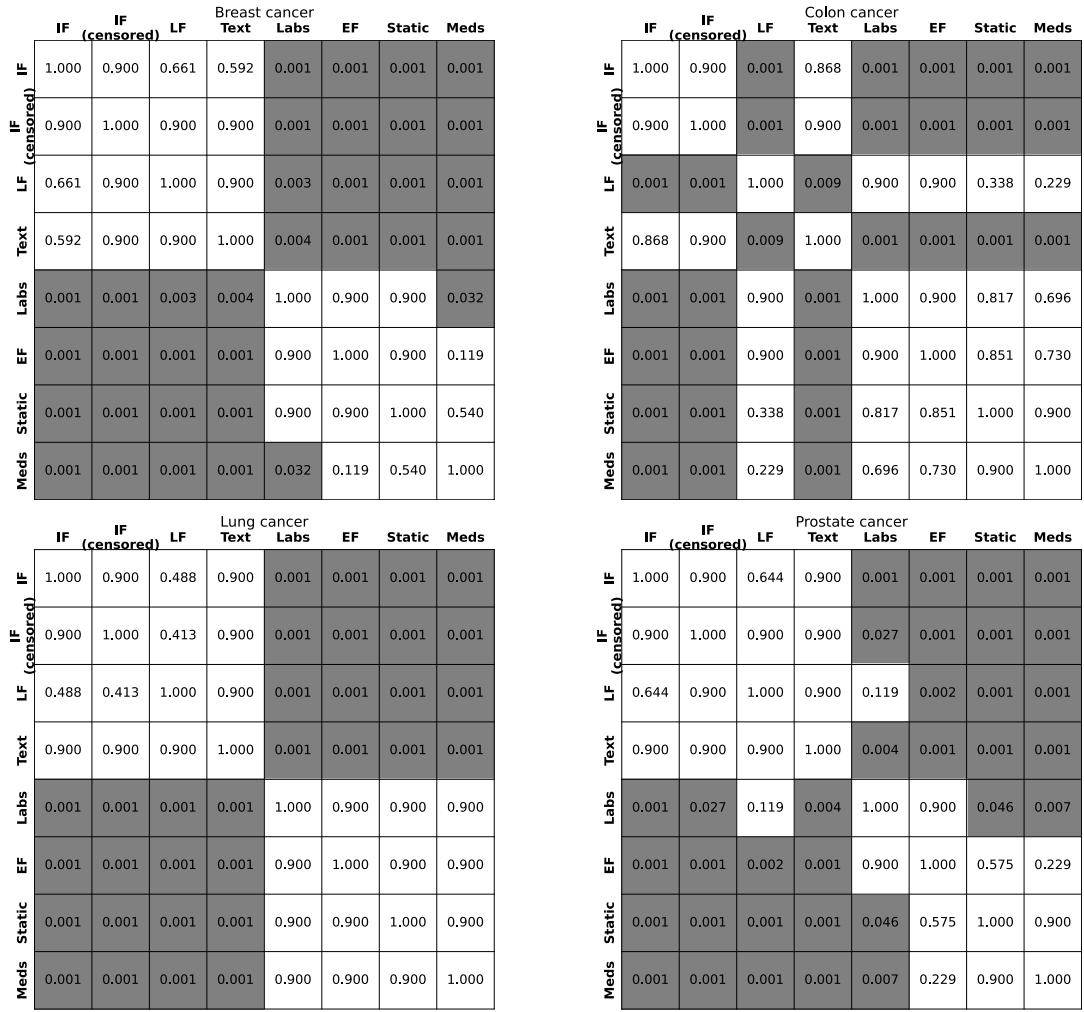

*Figure 3: The p-values resulting from the Nemenyi test, using the scikit-posthocs implementation, performed on the classification results of the best performing classifiers for each modality and combination of modalities, for all four cohorts. Greyed values denote significant difference at $p \leq 0.05$.*

Since this test uses the Studentized range distribution, equivalent to Tukey's Honest Significant Difference procedure, this approach controls the family-wise error rate (FWER), and the resulting p-values are already adjusted for the full set of $k(k-1)/2$ pairwise comparisons. Thus, all reported comparisons with $p \leq 0.05$ are significant, due to the FWER correction. The results clearly show that, for all four cohorts, the IF, IF (censored), and Text classifiers have a significantly different performance from the Labs, EF, Static, and Meds classifiers. In all cohorts apart from Colon cancer, this is also true for LF, highlighting the important contribution of clinical text to achieve reliable predictions, since it is only included in the data used to train these classifiers. In breast, lung, and prostate cancer, the IF, IF (censored), LF, and Text classifiers do not show any significant difference in performance among themselves. In all four cohorts, the EF classifier did not show significantly different performance compared to the classifiers of the structured modalities. In Colon cancer, there were no significant differences among the structured modalities' classifiers, the EF classifier, and the LF classifier. In all four cohorts, the IF (censored) classifier showed no significant differences compared to the IF classifier, demonstrating the robustness of our models to the absence of metastasis-specific cues in the clinical text.

We also report a critical difference diagram, to make the significant differences more apparent. The diagrams for each cohort can be seen in Figure 4.

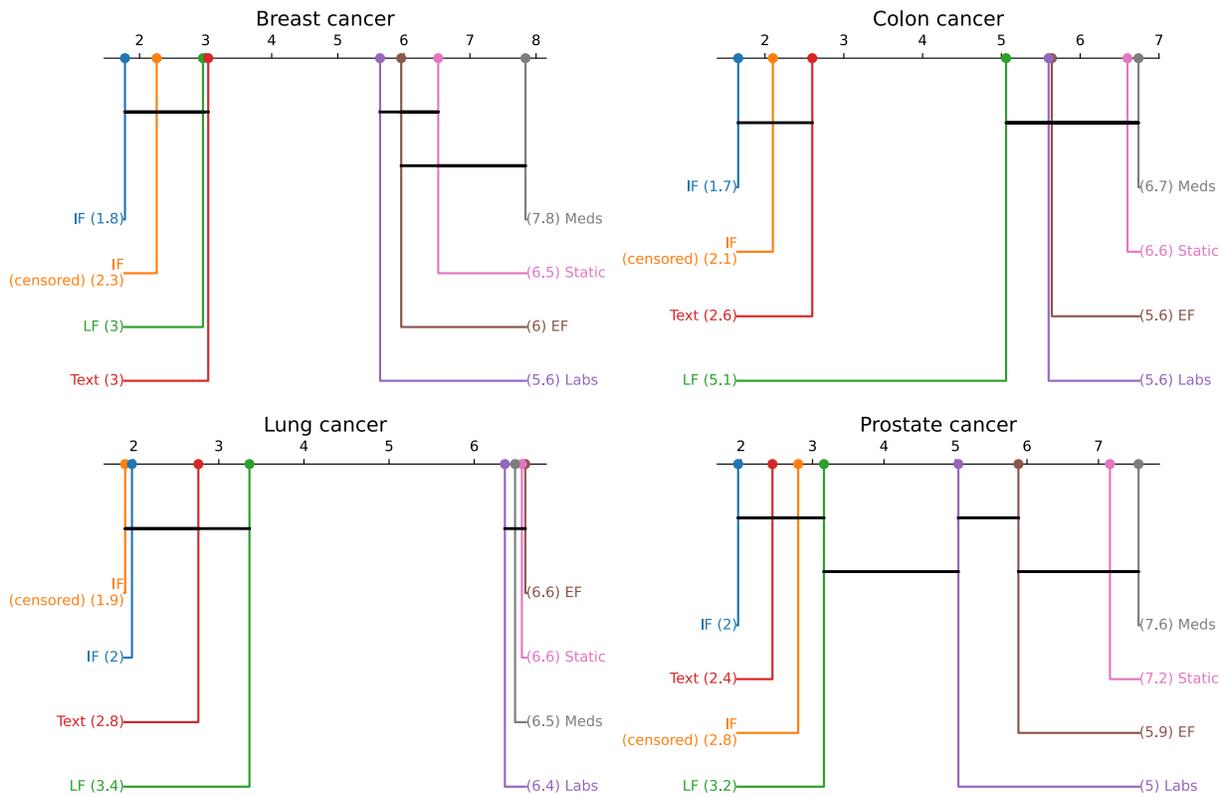

Figure 4: *The CD diagrams for each of the four cancer cohorts. The average rank for the best classifier in each modality, computed on the 5 test folds, is shown next to the name of the modality. A lower average rank indicates a better performance. Classifiers connected with a horizontal black line do not show a significant difference in their performance at $\alpha = 0.05$ under the Nemenyi post-hoc test, while larger rank separations than the CD indicate significant differences.*

# Appendix: Serialization of multimodal data for use with SHAP

To make our complex medical data compatible with SHAP, which only works with one-dimensional inputs, we first convert all values to a high-precision format (float64) to avoid any loss of detail. We then flatten the data into a single sequence while using special markers to retain its original structure. The value ∞ is inserted to separate different types of data (e.g., lab results vs. prescribed medications), and −∞ is used to mark boundaries between repeated measurements within each type (i.e., the values at the different points in the time series), as shown in Figure 5. This approach ensures that the data remains interpretable after transformation. It allows SHAP to provide meaningful explanations even with multimodal, time-series inputs.

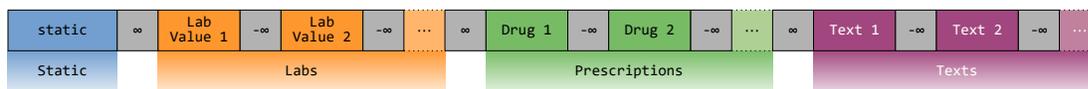

*Figure 5: Graphical representation of the serialization strategy used to adapt multimodal data for SHAP. Patient data from different modalities (Static, Labs, Meds, Text) are concatenated into a single sequence, with ∞ separating modalities and –∞ marking boundaries between repeated measurements.*

## Appendix: Analysis of SHAP faithfulness

To verify that our SHAP-based explanations are faithful to the model, we performed a perturbation experiment as also established in literature [47,48]. Specifically, we record model outputs while gradually masking the model input

1. from the **most to the least important feature** according to SHAP (*referred to as "High to Low"*)

2. from the **least to the most important feature** according to SHAP (*referred to as "Low to High"*)

3. in a random order (*referred to as "Random"*)

Intuitively, in *High to Low* order, the output should statistically change at low masked importance, as important information is withheld from the model. Accordingly, in *Low to High* order, the output should change class only at mostly perturbed inputs, as the most important information is masked last. In our experiments, we mask features with the respective modality's imputation symbol.

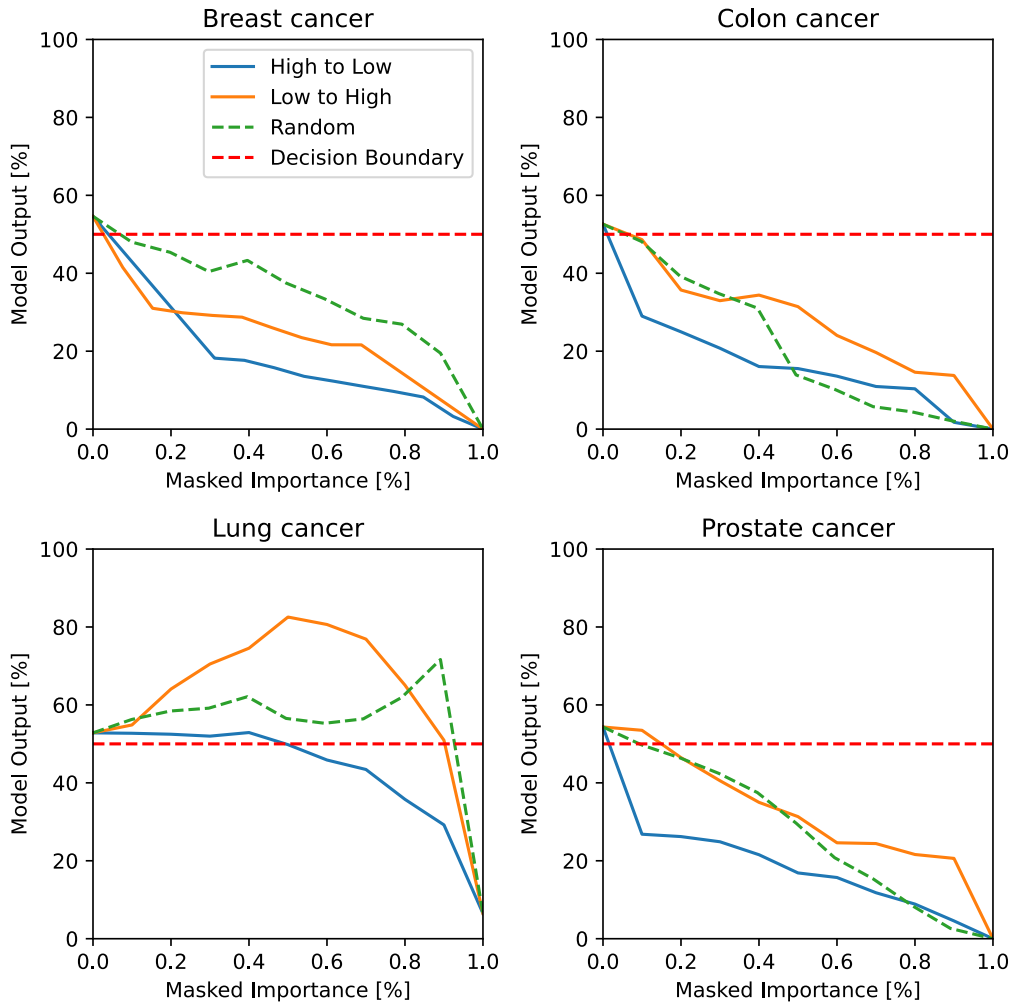

*Figure 6: Perturbation plots for the IF models in all cohorts.*

*Figure 6* shows the results of our experiments. For all models, we see an average model output of around 50% for completely unperturbed inputs (on the left of the x-axis) that eventually crosses the decision boundary for fully perturbed inputs (on the right of the x-axis). All the models show a slight bias towards the positive class (i.e.: presence of metastasis) when no perturbation is performed. The models trained on all cohorts, except the Lung cancer cohort, quickly cross the decision boundary, while the model trained on Lung cancer crosses the decision boundary only after a higher number of features is masked. In the models trained on the colon, lung, and prostate cancer cohorts, the rate of change for the model's decision is more pronounced with the "High to Low" strategy, while in the model trained on the breast cancer cohort, the rate of change is approximately the same.

In the breast cancer cohort, the "High to Low" and "Low to High" curves remain closely aligned across the perturbation range, both showing a gradual decline in model output.

This suggests that predictive information is more evenly distributed, with no small subset of features dominating the decision process. By contrast, in the lung cancer cohort, the "Low to High" curve initially rises before declining, indicating that predictions rely more heavily on a limited set of highly informative features. Masking these top-ranked features has a disproportionately strong effect, driving the model's decision towards a negative one, while removing less informative ones may even transiently sharpen the model's initial bias towards the positive class. All plots confirm at least moderate faithfulness, as the High to Low curve changes towards the perturbed class much earlier than, or, in the case of the model trained on the breast cancer cohort, extremely close to the Low to High curve.

## Appendix: Evaluation of Model Robustness After Text Censoring

Since Text showed disproportionately higher importance in the IF classifiers for lung and colon cancer, we sought to verify that the models were not relying on explicit textual cues such as direct mentions of 'metastasis.' To this end, we repeated the testing process after censoring sentences containing these terms.

Specifically, we removed all sentences containing the regular expression pattern "(\s[tnmTNM]\d|metastas)". The results of this analysis can be seen in the "IF (*censored*)" rows of Table 2. As shown in the table, in general the IF classifier maintained a substantially unvaried performance on the censored data, and only the performance of the classifier for colon cancer suffered from the censoring process, indicating a partial dependence on these sentences to produce correct predictions, while breast and lung cancer had a completely unchanged performance. Nonetheless, the IF model's F1 score on colon cancer remained above 70%, reflecting acceptable robustness. Moreover, since colon cancer is the smallest cohort, as indicated in Table 1, it is also the one which is more likely to suffer in terms of generalizability when using a single split evaluation procedure compared to using cross-validation, as shown in Table 4. These results suggest that our models are capable of identifying predictive markers even when clinical texts are stripped of indications of a clinician's suspicion of metastasis. By eliminating the sentences including these terms, the classifier can base its predictions on more relevant texts.

## Word count

The total word count for this paper is 8048 words.

Excluding references, footnotes, tables, the title page, this page, and appendices, the word count of the paper is 2995 words.